\documentclass[10pt,twocolumn,letterpaper]{article}

\usepackage{iccv}
\usepackage{times}
\usepackage{epsfig}
\usepackage{graphicx}
\usepackage{amsmath}
\usepackage{amssymb}
\usepackage[accsupp]{axessibility}

\def\ie{{\em i.e.}}
\def\eg{{\em e.g.}}
\def\etal{{\em et al.}}

\newcommand{\figref}[1]{Fig. \ref{#1}}
\newcommand{\tabref}[1]{Tab. \ref{#1}}

\newcommand{\secref}[1]{Section \ref{#1}}

\newcommand{\mc}[1]{\mathcal{#1}}

\newcommand{\br}[1]{\bm{\mathrm{#1}}}
\newcommand{\bs}[1]{\boldsymbol{\texttt{#1}}}

\usepackage{epstopdf}

\usepackage{ragged2e}
\usepackage{booktabs}
\usepackage{color}
\usepackage{multirow}
\usepackage{bm}
\usepackage{bbm}
\usepackage[misc]{ifsym}

\usepackage[pagebackref=true,breaklinks=true,letterpaper=true,colorlinks,bookmarks=false]{hyperref}

\iccvfinalcopy 

\def\iccvPaperID{2387} 

\ificcvfinal\fi

\begin{document}

\title{
Transformer-based Dual Relation Graph for Multi-label Image Recognition}
\author{Jiawei Zhao\textsuperscript{1}\thanks{Works done while interning at Tencent Youtu Lab.}\quad Ke Yan\textsuperscript{2}\quad Yifan Zhao\textsuperscript{1}\quad Xiaowei Guo\textsuperscript{2}\quad Feiyue Huang\textsuperscript{2}
\quad Jia Li\textsuperscript{1,3}\thanks{Jia Li is the Corresponding author.
URL: \url{http://cvteam.net}}\\
\textsuperscript{1}State Key Laboratory of Virtual Reality Technology and Systems, SCSE, Beihang University\\\
\textsuperscript{2}Tencent Youtu Lab, Shanghai, China\qquad
\textsuperscript{3}Peng Cheng Laboratory, Shenzhen, China\\
{\tt\small \textsuperscript{}\{zhaojiaweii, zhaoyf, jiali\}@buaa.edu.cn, \textsuperscript{} \{kerwinyan, scorpioguo, garyhuang\}@tencent.com}
}

\maketitle
\ificcvfinal\thispagestyle{empty}\fi

\begin{abstract}
    The simultaneous recognition of multiple objects in one image remains a challenging task, spanning multiple events in the recognition field such as various object scales, inconsistent appearances, and confused inter-class relationships. Recent research efforts mainly resort to the statistic label co-occurrences and linguistic word embedding to enhance the unclear semantics. Different from these researches, in this paper, we propose a novel Transformer-based Dual Relation learning framework, constructing complementary relationships by exploring two aspects of correlation,~\ie, structural relation graph and semantic relation graph. The structural relation graph aims to capture long-range correlations from object context, by developing a cross-scale transformer-based architecture. The semantic graph dynamically models the semantic meanings of image objects with explicit semantic-aware constraints. In addition, we also incorporate the learnt structural relationship into the semantic graph, constructing a joint relation graph for robust representations. With the collaborative learning of these two effective relation graphs, our approach achieves new state-of-the-art on two popular multi-label recognition benchmarks, \ie{} MS-COCO and VOC 2007 dataset.
\end{abstract}


\section{Introduction}
Multi-label image recognition aims at assigning multiple labels for multiple objects presented in one natural image. As a fundamental task in computer vision, multi-label image recognition can serve as prerequisites for many applications, such as weakly supervised localization and segmentation \cite{durand2017wildcat,ge2018multi,zhao2019BSANet}, attribute recognition \cite{liu2015deep,li2016human}
, scene understanding \cite{shao2015deeply}
and
recommendation systems \cite{yang2015pinterest,jain2016extreme}.
Benefiting from the development of deep learning techniques~\cite{he2016deep,vgg}, recent CNN-based architectures have made significant process in distinguishing multiple objects. But the accurate parsing of multi-label images still faces great challenges, including various object scales, inconsistent visual appearances, and confused inter-class relationships.

One intuitive solution for discovering visual consistency is to enhance the feature representation with self-attention mechanisms~\cite{zhu2017learning,wang2017multi,guo2019visual}. For example, Wang~\etal~\cite{wang2017multi} propose to automatically discover the attentional regions with a recurrent neural network, introducing discriminative features for representation learning. Beyond these improvements, Guo~\etal~\cite{guo2019visual} propose the assumption for visual perception consistency of attention regions and then amplify these regions by a visual consistency loss. Although the spatial representations of CNNs are strengthened by these techniques, the multi-label dependencies are not explicitly modeled, which is crucial for the understanding of multi-label relationships.

\begin{figure}
	\begin{center}
		\includegraphics[width=\linewidth]{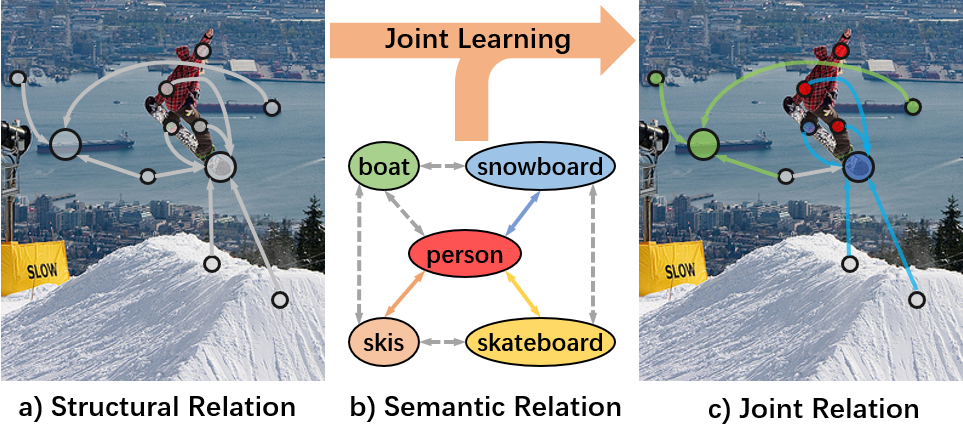}
		\caption{Motivation of the proposed dual relations. a) Structural relation provides long-term contextual relations for recognizing~\textit{snowboard}, while b) semantic relation builds dynamic correlations of co-occurred classes. These two relations jointly form a structural and semantic-aware image understanding.
		}\label{fig:motivation}
	\end{center}
\vspace{-0.3cm}
\end{figure}


To tackle this problem, recent ideas propose to learn the inter-class relationships based on the co-occurrences of multiple classes (\eg, the \textit{snowboard} should be attached with higher confidence if \textit{person} occurs). Pioneer works tend to model this relationship with Recurrent Neural Networks (RNN) \cite{wang2016cnn,chen2018order,yazici2020orderless}, while the co-occurred labels can be gradually refined in the sequential predictions. Inspired by the advanced improvements of Graph Convolutional Network (GCN) ~\cite{kipf2016semi}, tens of works~\cite{chen2019multi,chen2019learning,wang2020multi,ye2020attention,chen2021P_GCN,chen2020KGGR} propose to construct label-wise relationships based on the semantic meanings or statistical co-occurrences. For example, Chen~\etal~\cite{chen2019multi} propose to construct the graph model with the semantic word embeddings, forming a static label-wise relationship. However, this static relationship neglects the characteristics of each image, leading to negative optimization for objects with less-frequent co-occurrences. To solve this problem, some works~\cite{chen2019learning,ye2020attention,zhao2021GaRD} propose to construct dynamic graph based on the image-specific descriptors of high-level semantic features. Nevertheless, this modeling of multi-label relationship still shows its limitations: 1) the spatial interactions of contextual objects are not implicitly modeled in the
label-wise relation, 2) the features of high-level semantics are somewhat unstable and do not reflect specific semantic classes, 3) the representation of long-range context and various object scales are not considered.

To efficiently solve these deficiencies as well as the major challenges in multi-label image recognition, we propose to model a joint structural and semantic relationship of multi-label objects in one image. As in~\figref{fig:motivation}, considering the co-occurrence of semantic labels in vanilla class-wise relation models~\cite{chen2019multi}, absent classes would also be hallucinated (\ie, \textit{skis} and \textit{skateboard}). Beyond these demerits, the semantic meaning of one object should be not only decided by its intrinsic attributes but also the contextual information. In \figref{fig:motivation} a), the appearance of \textit{snowboard} is visually similar with \textit{skis} and \textit{skateboard}, and also shows high co-occurrence frequencies in \figref{fig:motivation} b). But human being can easily recognize it as a \textit{snowboard} based on the long-range contextual information (\textit{snow}) and even \textit{person} appearances.
Based on these investigations, we propose a collaborative framework with joint structural and semantic relational graphs in~\figref{fig:motivation} c), which depict the position-wise and class-wise relationship respectively.


To construct the structural graph, we make the first attempt to introduce the Transformer architecture~\cite{vaswani2017attention} into multi-label recognition.
This new attempt greatly broadens the receptive capability of conventional CNNs and draws position-wise long-term dependencies for object contextual correlations (\figref{fig:motivation} a)). Starting from this novel design, we further propose a cross-scale attention module to enhance the perceptive ability of various object scales. For the construction of semantic graph, we aim at constructing dynamic relation which is aware of object emergence and structural embedding. Different from previous works~\cite{chen2019learning,ye2020attention} with implicit high-level embedding, the graph nodes in our paper are explicitly constructed with semantic-aware constraints, reflecting features of specific classes. Beyond this explicit class-wise embedding, we incorporate the learnt structural graph embedding into the semantic relation construction from two aspects: adjacent correlation construction and feature-wise complementary. These two mechanisms efficiently endow the semantic graph with the perception of structural information, generating robust graph relations. With the collaborative learning of proposed structural and semantic relation, our proposed approach achieves state-of-the-art results on two most popular benchmarks,~\ie, MS-COCO~\cite{lin2014microsoft} and PASCAL VOC~\cite{everingham2010pascal}.

\begin{figure*}[htb]
	\begin{center}
		\includegraphics[width=1\textwidth]{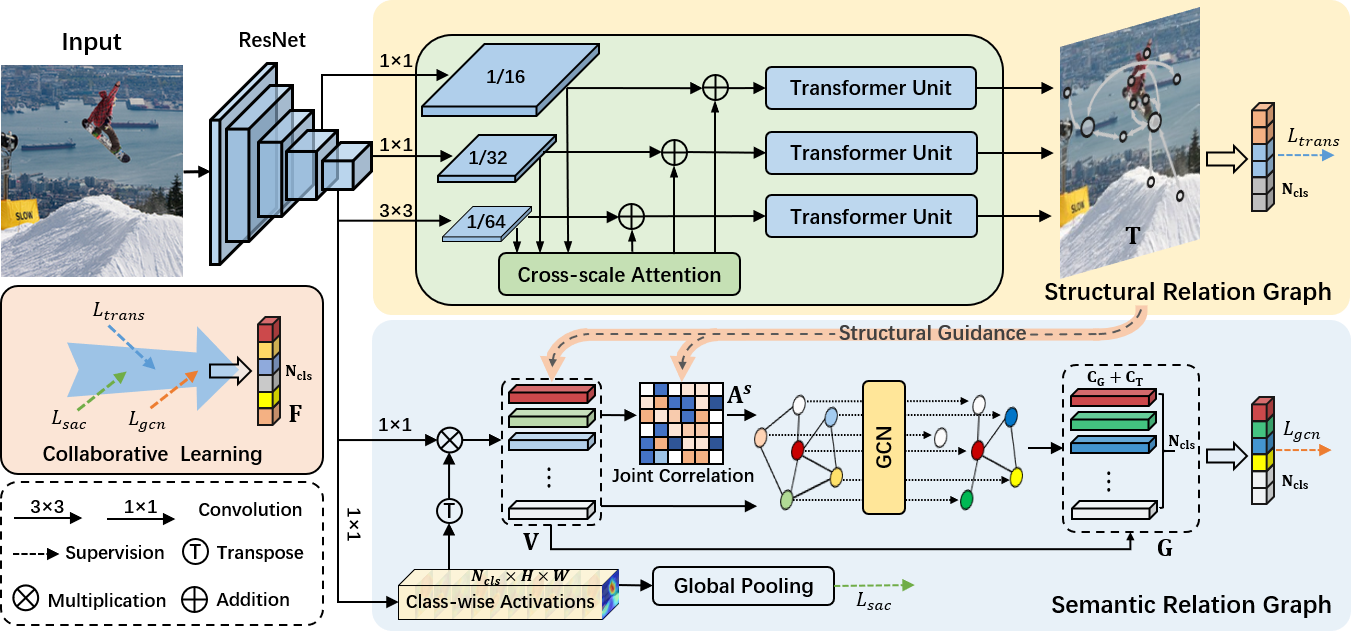}
		\caption{The overall architecture of our proposed Transformer-based Dual Relation Graph (TDRG) network, which consists of two essential modules: the structural relation graph module to incorporate long-term contextual information, the semantic relation graph module to model the dynamic class-wise dependencies.
		}\label{fig:pipeline}
	\end{center}
\vspace{-0.3cm}
\end{figure*}

In summary, our contribution is three-fold:
1) We propose a novel Transformer-based Dual Relation learning framework for multi-label image recognition tasks, which jointly models the structural and semantic information with Transformer architectures.
2) A transformer-based structural relation graph is constructed to incorporate long-term contextual information, building position-wise spatial relationships across different scales.
3) A semantic relation graph is constructed with explicit class-specific constraints and
structure-aware embedding, modeling the dynamic class-wise dependencies.

\section{Related Work}

\textbf{Multi-label Recognition.}
Recently multi-label recognition approaches mainly focus on two aspects, \ie{}, spatial information and co-occurrence dependency.

Some works \cite{wei2015hcp,yang2016exploit,wang2017multi,zhang2018multilabel,zhu2017learning,guo2019visual} devoted to exploit spatial information for improving recognition performance.
Previous pioneer works tend to coarsely locate multiple objects for recognition \cite{wei2015hcp,zhang2018multilabel,gao2021MCAR}. For example, Wei \etal{} \cite{wei2015hcp} generate multiple object proposals \cite{zitnick2014edge} and aggregate their label scores to obtain the final prediction. However, the performance of localization is unstable without additional annotations of proposals.
To solve this issue, recent works introduce attention mechanism to implicitly locate attentional regions and enhance the spatial representation \cite{zhu2017learning,wang2017multi,guo2019visual}. For example, Zhu \etal{} \cite{zhu2017learning} propose to capture spatial relationships between labels with self-attention mechanism. Wang \etal{} \cite{wang2017multi} utilize a proposal-free pipeline to iteratively locate attentional regions and capture their contextual dependencies.

Some other works \cite{wang2016cnn,liu2017semantic,chen2018recurrent,chen2018order,yazici2020orderless} devoted to build co-occurrence dependency with Recurrent Neural Network (RNN) \cite{hochreiter1997long}.
For example, Wang \etal{} \cite{wang2016cnn} combine RNNs with CNNs to capture semantic label dependency and predict labels with a predefined order.
Chen \etal{} \cite{chen2018order} design an order-free RNN to avoid propagating prediction errors during inference process.
However, these RNN-based methods explore limited relationships between labels in a sequential manner,  hence recent works introduce Graph Convolutional Network (GCN) \cite{kipf2016semi} to fully exploit pair-wise relationships \cite{chen2019multi,chen2019learning,wang2020multi,ye2020attention,you2020cross,xu2020joint,chen2019DER,chen2021P_GCN,chen2020KGGR}. For example, Chen \etal{} \cite{chen2019multi}
propose a directed graph over word embedding of labels to model the label correlations.
Chen \etal{} \cite{chen2019learning} build a semantic-specific graph to incorporate high-level features into word embeddings for better semantic-specific features and explore their interactions.


\textbf{Relationship Modeling.}
Different from CNNs and RNNs, transformer is recently proposed to extract intrinsic features with self-attention mechanism \cite{parikh2016decomposable}. Transformer has demonstrated its success in natural language processing tasks \cite{vaswani2017attention,devlin2018bert}. As the pioneer work, Vaswani \etal{} \cite{vaswani2017attention} first propose the vanilla Transformer architecture, which is based on self-attention mechanisms for machine translation. Transformer has not only obtained great breakthrough in NLP tasks, but also shows huge potential in Computer Vision (CV) tasks \cite{dosovitskiy2020image,carion2020end,zhu2020deformable,yang2020learning,lanchantin2021CTran}. For example, recently, Dosovitskiy \etal{} \cite{dosovitskiy2020image} propose a pure transformer architecture on sequential image patches for image recognition task. Carion \etal{} \cite{carion2020end} design a fully end-to-end object DEtection TRansformer (DETR), which shows impressive performance on object detection. Zhu \etal{} \cite{zhu2020deformable} introduce a deformable attention module to solve the defects of DETR, \eg{}, poor performance on small objects.
However, as an effective architecture for relationship modeling, transformer is less explored in multi-label recognition tasks.

\section{Approach}
In this section, we introduce a novel collaborative learning framework with joint structural and semantic relational graphs for multi-label recognition, depicting the position-wise and class-wise dependencies respectively in \figref{fig:pipeline}. The first structural relation graph aims to capture long-term contextual information and build spatial relationships across different scales in Section \ref{Position-wise Relation}. In Section \ref{Class-wise Relation}, a semantic relation graph is proposed to exploit dynamic co-occurrence dependencies with structure-aware embedding. In the end, we joint structural and semantic relations in a collaborative learning manner in Section \ref{Learning Objective}.


Given an input image $\mathcal{I}$, let \{$\br{X}_1, \cdots, \br{X}_s$\} $=\Phi_{S}(\mathcal{I})$ be the multi-scale features encoded by the backbone network $\Phi_{S}$ with a channel-reduction transformation,~\eg, $1\times1$ and $3\times3$ convolution.
To construct the structural relation graph nodes $\br{T}$, we introduce $s$ transformer units $\mc{G}^{trans}$ to capture long-term contextual information and build position-wise relationships with a cross-scale attention module $\Psi_i(\cdot)$:
 \begin{equation} \label{eq:structural pipeline}
 \br{T} =  \mathop{\bs{concat}}\limits_{i=1}^{s} (\mc{G}^{trans}_i(\Psi_i(\br{X}_i;\{\br{X}\}_{k=1}^{s}))) \in \mathbb{R}^{N_T \times C_T},
 \end{equation}
where $N_T$ and $C_T$ denote the number and dimension of structural relation nodes $\br{T}$ respectively. To construct the nodes of semantic relation graph $\br{G}$, we model dynamic class-wise dependencies with explicit semantic-aware constraints and structural guidance:
\begin{equation} \label{eq:semantic pipeline}
\br{G} = \mc{G}^{sem}((\mc{C}(\br{\br{X}}), \br{T});\mc{A}(\br{T},\mc{C}(\br{\br{X}}))) \in \mathbb{R}^{N_{cls}\times (C_G+C_T)},
 \end{equation}
where $\mc{G}^{sem}$ denotes the semantic graph neural network, $\mc{C}(\cdot)$ denotes the semantic-aware constraints, $\mc{A}(\cdot)$ denotes the joint relational correlation matrices of $\mc{G}^{sem}$, $N_{cls}$ and $C_G$ denote the number of categories and the dimension of semantic-specific vectors.
With these two complementary relation graphs, we further conduct a collaborative learning manner to get the final prediction $\br{F}$:
\begin{equation} \label{eq:fusion pipeline}
\br{F} = \psi_t(\bs{GMP}(\br{T}))\biguplus \psi_g(\br{G}) \in \mathbb{R}^{N_{cls}},
 \end{equation}
where $\psi_{\{t, g\}}$ denote the category classifier for structural and semantic relation graph respectively, $\bs{GMP}(\cdot)$ denotes the global max-pooling operation, $\biguplus$ denotes the weighted fusion operation of two relation graphs.


\subsection{Structural Relation Graph}
\label{Position-wise Relation}
As aforementioned, one crucial problem in multi-label recognition is to capture long-term contextual information and build structural interactions between different objects.
Due to the intrinsic flaws of CNN-based architectures, the position correlations are locally conducted without perceiving global contextual information. To mitigate this issue, we make the first attempt to introduce Transformer into multi-label recognition tasks to capture long-term contextual information and build position-wise spatial relationships.

\textbf{Revisiting Transformer for Structural Relation.}
In the field of natural language processing, the conventional transformers~\cite{vaswani2017attention} take language sentences as input and build relationships between different semantic words from global perspectives.
Different from language processing, images cannot be directly converted into sequence form. Hence, there are two popular ways to apply transformer on images, embedding transformer into CNN backbones \cite{carion2020end} and applying transformer on sequential embedding features of image patches \cite{dosovitskiy2020image}. The latter leads to a high computation burden for network optimization with limited data.
Hence in our framework, we adopt the former scheme to capture global contextual information.



\begin{figure}
	\begin{center}
		\includegraphics[width=\linewidth]{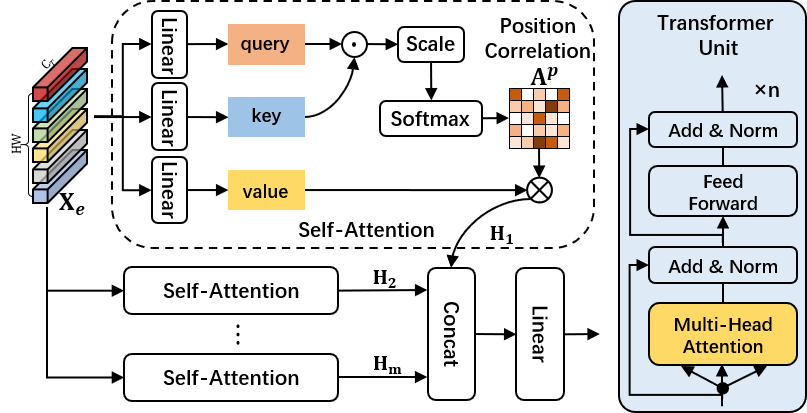}
		\caption{Illustration of transformer unit and multi-head self-attention module. Each head provides structural self-attention mechanism by multiplying query, key and value features.
		}\label{fig:MHSA}
	\end{center}
\vspace{-0.3cm}
\end{figure}

We apply standard transformer encoder structure as transformer unit $\mc{G}^{trans}_i$ to build long-term relationships between pair-wise positions. As shown in \figref{fig:MHSA}, each transformer unit consists of $n$ groups multi-head self-attention modules and feed forward networks, which is composed of two linear transformation layers. The detailed structure of multi-head self-attention module is illustrated in \figref{fig:MHSA}. For each head, we first adopt relative positional encoding $\mc{E}(\cdot)$ on the channel-wise transformed feature $\phi(\br{X})$ to keep the position information:
 \begin{equation} \label{eq:positional}
\br{X}_e = \mc{R}(\phi(\br{X}))+\mc{E}(\mc{R}(\phi(\br{X})))\in \mathbb{R}^{HW\times C_T},
 \end{equation}
where $\mc{R}(\cdot)$ denotes the reshape operation, which squeezes the spatial dimensions into one dimension.
Then we respectively obtain query, key and value projections of encoded features by linear transformation layers. To build and enhance global position relationships, we calculate the positional correlation matrix $\br{A}^p$ by query and key, and reweight value with $\br{A}^p$ by multiplication:
 \begin{equation} \label{eq:positional correlation matrix}
 \begin{gathered}
 \br{A}^p = \bs{softmax}(\frac{\br{X}_e\br{W}_Q (\br{X}_e\br{W}_K)^{\top}}{\sqrt{C_T}}), \\
  \br{H} = \br{A}^p \br{X}_e \br{W}_V,
  \end{gathered}
 \end{equation}
where $\br{W}_{\{Q,K,V\}}$ are the learnable weights of query, key and value projections, $\br{H}$ is the enhanced feature of one head.
Different heads could mine different structural relationships due to different projections. Hence employing multiple heads could capture more comprehensive structural relationships to enrich the representations. For multi heads, we concatenate results from multi heads and fuse them with a linear transformation layer.

\textbf{Cross-scale Attention Transformers.}
Small objects tend to have lower performance in multi-label recognition due to the position information of small objects may lost in low-resolution features especially for challenging datasets, \eg{}, MS-COCO.
To address this issue, a natural idea is to consider more high-resolution features to retain the position information of small objects. In fact, high-resolution features indeed improve the performance of small objects, but also introduce more computation burden and noise, which lower the performance of other objects. Towards this concern, we propose a simple yet effective cross-scale attention module as a trade-off between performance and computation cost, which effectively improves the capacity of our structural relation graph.


To suppress noise between different scales and enhance the structure information of small objects, we propose cross-attention feature fusion strategy $\Psi_i(\cdot)$. We extract the common positions while alleviating the ambiguous ones by position-wise multiplication operation after up-sampling features of different scales. To enhance the positional information, the extracted feature is respectively down-sampled and enhanced with position-wise addition operation. Thus the structural feature $\br{T}_i$ of $i$th scale can be formed by this serialized operation:
 \begin{equation} \label{eq:multi-scale cross attention}
\br{T}_i =\mc{G}^{trans}_i(\mc{D}(\prod_{i}^{s} \mc{U}(\br{X}_i)) + \br{X}_i),
 \end{equation}
where $\mc{U}(\cdot)$ and $\mc{D}(\cdot)$ denote up- and down-sampling operation.
Then we input enhanced features into weight-sharing transformer unit to respectively capture the structural relationships with different scales, and the final feature $\br{T}$ in Eq.~\eqref{eq:structural pipeline} can be obtained by concatenating each $\br{T}_i$.

\subsection{Semantic Relation Graph}\label{Class-wise Relation}
\begin{figure}
	\begin{center}
		\includegraphics[width=\linewidth]{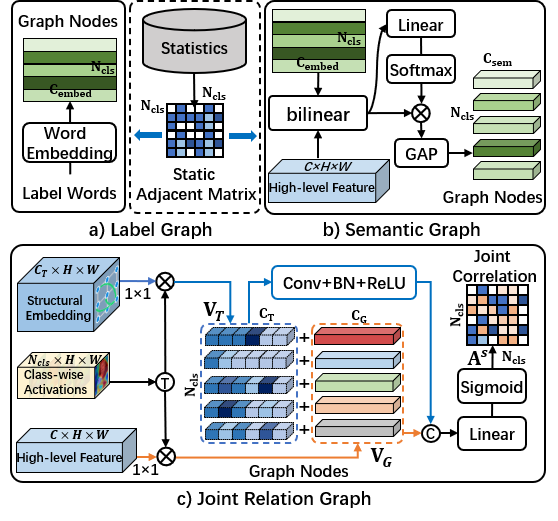}
		\caption{Illustrations of three typical graph constructions. a) Label graph \cite{chen2019multi}: building graph based on the statistical priors of label co-occurrences. b) Semantic graph \cite{chen2019learning}: incorporating high-level features beyond word embeddings. c) Our joint relation graph: building graph nodes by joint structural embedding and semantic-aware constraints and dynamically constructing the correlation matrix in a learnable manner.
		}\label{fig:ThreeGraphs}
	\end{center}
\vspace{-0.3cm}
\end{figure}

\textbf{Motivation and Discussions.} Motivated by the co-occurrence dependencies in multi-label learning, existing works usually resort to graph networks to model this relationship into deep CNNs. Pioneer works~\cite{chen2019multi} in~\figref{fig:ThreeGraphs} a) tend to build static correlations of different linguistic word embedding from statistic priors. However, in this label graph, the characteristic of each image are less taken into consideration, which would lead to the hallucination of nonexistent objects and suppression of less common co-occurrences. To this end, in~\figref{fig:ThreeGraphs} b), semantic graphs in~\cite{chen2019learning} are built to incorporate high-level features into word embedding. Despite the additional dependencies on word embedding and dataset statistics, high-level features only present \textit{implicit} semantics for graph construction and detailed relationship between multiple objects are still neglected.

To revisit the graph construction of multi-label learning, here we propose a joint relation graph in~\figref{fig:ThreeGraphs} c), involving two meaningful cues for object relation discovery: 1) introducing \textit{explicit} semantic-aware high-level features via an auxiliary semantic-aware constraint; 2) incorporating structural relationship for graph nodes as well as correlations. The former cue helps to construct an \textit{explicit} relationship of semantic classes while the latter cue provides the graph a spatial awareness of contextual objects.




\textbf{Semantic-aware Constraints.}
Different from previous researches using high-level features $\br{X}$ for implicit semantic, here we introduce the explicit embedding with class-specific vectors $\br{M} = \phi_m(\br{X}) \in \mathbb{R}^{N_{cls}\times H\times W}$, which is regularized by explicit classification constraints. $\phi_m$ denotes the learnable $1\times 1$ convolutional layer. Hence we conduct a high-order fusion to form the semantic-aware features $\br{V}_G$:
 \begin{equation} \label{eq:semantic-aware constraints}
\br{V}_G = \mc{R}(\br{M}) \phi_g(\mc{R}(\br{X})^{\top}) \in \mathbb{R}^{N_{cls}\times C_G},
 \end{equation}
 where $\phi_g(\cdot)$ denotes a dimension-reduction operation from $C$ to $C_G$.  However, how to ensure the representation quality of $\br{V}_G$ for each class is less explored, which affects the subsequent modeling process as an important precondition. To solve this issue, we adopt a global pooling operation, \ie{} top-k max-pooling on $\br{M}$ to squeeze spatial dimensions and then apply an auxiliary loss on $\bs{KMP}(\br{M}) \in \mathbb{R}^{N_{cls}}$ to constraint $\br{M}$ for learning more accurate initial activation maps and less noise for each class.



Besides the semantic-aware vectors, we also introduce structural information from structural relation graphs to incorporate long-term contextual information and position-wise relationships:
 \begin{equation} \label{eq:structural guidance}
\br{V}_T = \mc{R}(\bs{GAP}(\br{T}))\in \mathbb{R}^{N_{cls}\times C_T},
 \end{equation}
where $\br{V}_T$ denotes the structure-aware vectors and $\bs{GAP}$ denotes the global average pooling operation.

\begin{table*}[]
\centering{
\setlength\tabcolsep{4pt}

	\caption{Comparisons with state-of-the-art methods on the MS-COCO dataset. The performance of our approach based on three resolution settings are reported. $R_{train}$ and $R_{test}$ denote resolution used in training and testing stage. * denotes the performance of our implementation.}
	\label{table: COCO benchmark}
\begin{tabular}{c|c|c|cccccc|cccccc}
\hline
\multirow{2}{*}{Methods} & \multirow{2}{*}{($R_{train}, R_{test}$)} & \multirow{2}{*}{mAP} & \multicolumn{6}{c|}{All}  & \multicolumn{6}{c}{Top 3}
\\
&        &  & CP   & CR   & CF1  & OP   & OR   & OF1  & CP   & CR   & CF1  & OP   & OR   & OF1  \\ \hline \hline
CNN-RNN \cite{wang2016cnn}              & $(-,-)$ & 61.2          & -    & -    & -    & -    & -    & -    & 66.0 & 55.6 & 60.4 & 69.2 & 66.4 & 67.8 \\
RNN-Attention \cite{wang2017multi}         &$(-,-)$   & -                    & -    & -    & -    & -    & -    & -    & 79.1 & 58.7 & 67.4 & 84.0 & 63.0 & 72.0 \\
Order-Free RNN \cite{chen2018order}         & $(-,-)$ & -                    & -    & -    & -    & -    & -    & -    & 71.6 & 54.8 & 62.1 & 74.2 & 62.2 & 67.7 \\
SRN \cite{zhu2017learning}                   & $(224, 224)$  & 77.1                 & 81.6 & 65.4 & 71.2 & 82.7 & 69.9 & 75.8 & 85.2 & 58.8 & 67.4 & 87.4 & 62.5 & 72.9 \\
PLA  \cite{yazici2020orderless}            & $(288,288)$     & -                 & 80.4 & 68.9 & 74.2 & 81.5 & 73.3 & 77.2 & -    & -    & -    & -    & -    & -    \\
ResNet-101 *\cite{he2016deep}              &$(448,448)$  & 78.6                 & 82.4 & 65.5 & 73.0 & 86.0 & 70.4 & 77.4 & 85.9 & 58.6 & 69.7 & 90.5 & 62.8 & 74.1 \\

ML-GCN \cite{chen2019multi}                 & $(448,448)$ & 83.0                 & 85.1 & 72.0 & 78.0 & 85.8 & 75.4 & 80.3 & 89.2 & 64.1 & 74.6 & 90.5 & 66.5 & 76.7 \\
KSSNet \cite{wang2020multi}                 &$(448,448)$  & 83.7                 & 84.6 & \textbf{73.2} & 77.2 & \textbf{87.8} & 76.2 & \textbf{81.5} & -    & -    & -    & -    & -    & -    \\

Ours&  $(448,448)$   & \textbf{84.6}  & \textbf{86.0} & 73.1 & \textbf{79.0} & 86.6 & \textbf{76.4} & 81.2 & \textbf{89.9} & \textbf{64.4} & \textbf{75.0} & \textbf{91.2} & \textbf{67.0} & \textbf{77.2} \\ \hline \hline
ADD-GCN  \cite{ye2020attention}        &$(448,576)$ & 85.2                 & 84.7 & \textbf{75.9} & \textbf{80.1} & 84.9 & \textbf{79.4} & 82.0 & 88.8 & \textbf{66.2} & \textbf{75.8} & 90.3 & \textbf{68.5} & \textbf{77.9} \\
Ours   &   $(448,576)$   & \textbf{85.8}        & \textbf{87.9} & 73.6 & \textbf{80.1} & \textbf{87.9} & 77.3 & \textbf{82.3} & \textbf{91.3} & 64.8 & \textbf{75.8} & \textbf{92.0} & 67.6 & \textbf{77.9} \\ \hline \hline
SSGRL \cite{chen2019learning}           &$(576,576)$ & 83.8                 & \textbf{89.9} & 68.5 & 76.8 & \textbf{91.3} & 70.8 & 79.7 & \textbf{91.9} & 62.5 & 72.7 & \textbf{93.8} & 64.1 & 76.2 \\
C-Tran \cite{lanchantin2021CTran}           &$(576,576)$ & 85.1                 & 86.3 & 74.3 & 79.9 & 87.7 & 76.5 & 81.7 & 90.1 & \textbf{65.7} & 76.0 & 92.1 & \textbf{71.4} & 77.6 \\
Ours   & $(576,576)$ &   \textbf{86.0}    &    87.0  & \textbf{74.7}  & \textbf{80.4} & 87.5 & \textbf{77.9} & \textbf{82.4} & 90.7 & 65.6 & \textbf{76.2} & 91.9 & 68.0 & \textbf{78.1}     \\ \hline
\end{tabular}
}
\end{table*}




\textbf{Joint Relation Graph.}
Graph neural networks propagate messages between adjacent nodes based on correlation matrix.
As in~\figref{fig:ThreeGraphs} c), we build the joint correlation matrix $\br{A}^{s}$ from two aspects, \ie{} semantic correlations $\br{V}_G$ and structural correlations $\br{V}_T$ in a learnable manner:
 \begin{equation} \label{eq:joint correlation construction}
\br{A}^{s} \! = \! \bs{sigmoid}(\varphi_c(\bs{concat}(\varphi_t(\br{V}_T), \br{V}_G))) \in \mathbb{R}^{N_{cls}\times N_{cls}},
 \end{equation}
where $\varphi_{\{c,t\}}$ denote the learnable dimension transformation operation, \eg{}, $1\times 1$ convolutional layer.



Obtaining graph nodes $\br{V}$ = $\bs{concat}(\br{V}_G, \br{V}_T)$ and correlation matrix $\br{A}^{s}$, we further model the joint co-occurrence dependencies between joint structural and semantic-aware vectors based on correlation matrix using Kipf \etal{}'s \cite{kipf2016semi} Graph Convolutional Networks, which can be formulated as:
 \begin{equation} \label{eq:GCN}
\br{G} = \delta(\br{A}^{s} \br{V} \br{W}_G) + \br{V} \in \mathbb{R}^{N_{cls}\times (C_G+C_T)},
 \end{equation}
where $\br{G}$ denotes the updated semantic relation graph, $\br{W}_G \in \mathbb{R}^{(C_G+C_T) \times (C_G+C_T)}$ is the learnable graph weights. $\delta(\cdot)$ denotes the LeakyReLU \cite{maas2013rectifier} activation function.

\subsection{Learning Objective}
\label{Learning Objective}
With structural and semantic relation graphs obtained, we further joint their predictions for training in a collaborative learning manner (see \figref{fig:pipeline}). We adopt $\mc{L}_{sac}$ to constraint semantic-aware features in~\secref{Class-wise Relation}. To accelerate the convergence process, we adopt $\mc{L}_{trans}$ and $\mc{L}_{gcn}$ on the prediction result of structural and semantic relation graphs respectively. Besides, we adopt $\mc{L}_{joint}$ for the final prediction result. All these loss functions are supervised with typical multi-label classification entropy $\mc{L}_{joint}$ = $-\sum_{i=1}^{N_{cls}}\br{y}_{i}\log(\br{p}_{i}), \br{p}, \br{y} \in \mathbb{R}^{N_{cls}}, \br{y} \in \{0,1\}$. Hence the final learning objective $\mc{L}_{sum}$ can be formulated as:
  \begin{equation} \label{eq:BCE}
\mc{L}_{sum} = \mc{L}_{joint} + \mc{L}_{sac}+ \mc{L}_{trans}+ \mc{L}_{gcn}.
 \end{equation}
With this collaborative regularization, the final classification embedding in Eq.~\eqref{eq:fusion pipeline} can jointly be aware of structural and semantic information for multi-label understanding.



\begin{table*}[t]
\centering{
\setlength\tabcolsep{3pt}

	\caption{Comparisons with state-of-the-art methods on the VOC 2007 dataset. The performance of our approach based on resolution $448\times 448$ is reported. * denotes the performance of our implementation.}
	\label{table: VOC2007 benchmark}
	\resizebox{\textwidth}{!}{
\begin{tabular}{c|c|c|c|c|c|c|c|c|c|c|c|c|c|c|c|c|c|c|c|c||c}
\hline
Methods &
  aero &
  bike &
  bird &
  boat &
  bottle &
  bus &
  car &
  cat &
  chair &
  cow &
  table &
  dog &
  horse &
  motor &
  person &
  plant &
  sheep &
  sofa &
  train &
  tv &
  mAP \\ \hline \hline
CNN-RNN \cite{wang2016cnn} & 96.7 & 83.1 & 94.2 & 92.8 & 61.2 & 82.1 & 89.1 & 94.2 & 64.2 & 83.6 & 70.0 & 92.4 & 91.7 & 84.2 & 93.7 & 59.8 & 93.2 & 75.3 & \textbf{99.7} & 78.6 & 84.0 \\
RNN-Attention \cite{wang2017multi} &
  98.6 &
  97.4 &
  96.3 &
  96.2 &
  75.2 &
  92.4 &
  96.5 &
  97.1 &
  76.5 &
  92.0 &
  87.7 &
  96.8 &
  97.5 &
  93.8 &
  98.5 &
  81.6 &
  93.7 &
  82.8 &
  98.6 &
  89.3 &
  91.9 \\
Fev+Lv \cite{yang2016exploit} & 97.9 & 97.0 & 96.6 & 94.6 & 73.6 & 93.9 & 96.5 & 95.5 & 73.7 & 90.3 & 82.8 & 95.4 & 97.7 & 95.9 & 98.6 & 77.6 & 88.7 & 78.0 & 98.3 & 89.0 & 90.6 \\
Atten-Reinforce \cite{chen2018recurrent} &
  98.6 &
  97.1 &
  97.1 &
  95.5 &
  75.6 &
  92.8 &
  96.8 &
  97.3 &
  78.3 &
  92.2 &
  87.6 &
  96.9 &
  96.5 &
  93.6 &
  98.5 &
  81.6 &
  93.1 &
  83.2 &
  98.5 &
  89.3 &
  92.0 \\
  ResNet-101 * \cite{he2016deep} &
  99.8 &
  98.3 &
  98.0 &
  98.0 &
  79.5 &
  93.2 &
  96.8 &
  97.7 &
  79.9 &
  91.0 &
  86.6 &
  98.2 &
  97.8 &
  96.4 &
  98.8 &
  79.4 &
  94.6 &
  82.9 &
  99.1 &
  92.1 &
  92.9 \\
SSGRL \cite{chen2019learning}  & 99.5 & 97.1 & 97.6 & 97.8 & \textbf{82.6} & 94.8 & 96.7 & 98.1 & 78.0 & \textbf{97.0} & 85.6 & 97.8 & 98.3 & 96.4 & 98.1 & 84.9 & 96.5 & 79.8 & 98.4 & 92.8 & 93.4 \\
ML-GCN \cite{chen2019multi} & 99.5 & 98.5 & \textbf{98.6} & 98.1 & 80.8 & 94.6 & 97.2 & \textbf{98.2} & 82.3 & 95.7 & 86.4 & 98.2 & 98.4 & 96.7 & 99.0 & 84.7 & 96.7 & 84.3 & 98.9 & 93.7 & 94.0 \\
ADD-GCN *\cite{ye2020attention} & 99.7 & 98.5 & 97.6 & 98.4 & 80.6 & 94.1 & 96.6 & 98.1 & 80.4 & 94.9 & 85.7 & 97.9 & 97.9 & 96.4 & 99.0 & 80.2 & 97.3 & 85.3 & 98.9 & 94.1 & 93.6 \\ \hline \hline
Ours    & \textbf{99.9} & \textbf{98.9} & 98.4 & \textbf{98.7} & 81.9 & \textbf{95.8} & \textbf{97.8} & 98.0 & \textbf{85.2} & 95.6 & \textbf{89.5} & \textbf{98.8} & \textbf{98.6} & \textbf{97.1} & \textbf{99.1} & \textbf{86.2} & \textbf{97.7} & \textbf{87.2} & 99.1 & \textbf{95.3} & \textbf{95.0} \\ \hline
\end{tabular}
}
}
\end{table*}

\section{Experiments}

\subsection{Datasets and Evaluation Metrics}

\textbf{MS-COCO Benchmark.}
Microsoft COCO \cite{lin2014microsoft} is a widely-used benchmark for many vision tasks, \eg{} object detection, segmentation and multi-label recognition.
It contains 82,081 images in train set and 40,137 images in validation set from 80 common object categories. On average, each image has 2.9 labels. Especially, it contains a large number of small objects, which is more challenging for multi-label recognition. Following \cite{chen2019multi,ye2020attention,chen2019learning}, we evaluate the performance of all methods on the validation set.

\textbf{VOC 2007 Benchmark.}
PASCAL VOC 2007
\cite{everingham2010pascal} is another popular benchmark for multi-label recognition.
It contains 5,011 images in train and validation set and 4,952 images in test set from 20 common object categories. On average, each image has 1.4 labels. Following \cite{chen2019multi}, we train our approach on the trainval set and evaluate on the test set.

\textbf{Evaluation Metrics.}
To quantitatively evaluate the performance of our approach and state-of-the-art methods, we adopt the average per-class precision (CP), recall (CR), F1 (CF1), the average overall precision (OP), recall (OR), F1 (OF1) and mean average precision (mAP) as evaluation metrics. For fair comparisons, we also report top-3 results. Notably, precision/recall/F1-score may be affected by the threshold, which is set as 0.5 in our setting. Among all metrics, AP and mAP are the most important metrics which can provide a more comprehensive comparison.

\subsection{Implementation Details}
We adopt ResNet-101 \cite{he2016deep} pre-trained on ImageNet \cite{deng2009imagenet} as our backbone. In the training stage, input images are first resized into 512$\times$512, then random cropped and resized into 448 $\times$ 448 with random horizontal flips for augmentation. In the testing stage, input images are resized into 448 $\times$ 448. We use the SGD optimizer with momentum of 0.9 and weight decay of 1e-4. The initial learning rate is 0.01 for VOC 2007 and 0.03 for MS-COCO, which decays by a factor of 10 for every 30 epochs. The batch size is set as 16 for VOC 2007 and 32 for MS-COCO on each GPU. The network converge quickly and only needs 50 epochs in total for training.  Detailed experiments on hyper-parameters can be found in supplementary materials.
We set hidden dimension $C_T=512$ and $C_G=512$. The transformer unit consists of 3 layers and each layer has 4 attention heads. The semantic graph neural network has one layer. As mentioned in Eq. \eqref{eq:fusion pipeline}, We apply a weight coefficient $\alpha$ on structural relation graph and $(1-\alpha)$ on semantic relation graph. We set $\alpha=0.7$ to achieve the best performance.
All experiments are conducted on two NVIDIA Tesla V100 GPUs.

\begin{table}[t]
\centering{
\setlength\tabcolsep{4pt}
\caption{Ablation study for different components. $\mc{R}_{Structural} $ and $\mc{R}_{Semantic}$ denote structural and semantic relation graphs. $\mc{M}_{Trans}$ denotes transformer units. $\mc{M}_{CSA}$ denotes cross-scale attention module. $\mc{M}_{GCN}$ denotes graph convolutional network. $\mc{R}_{SAC}$ denotes semantic-aware constraints. }
\label{table:ablation}
\begin{tabular}{cc|cc|cc}
\hline
\multicolumn{2}{c|}{$\mc{R}_{Structural}$} & \multicolumn{2}{c|}{$\mc{R}_{Semantic}$} & \multicolumn{2}{c}{mAP} \\

$\mc{M}_{Trans}$ & $\mc{M}_{CSA}$ & $\mc{M}_{GCN}$ & $\mc{M}_{SAC}$ & COCO    & VOC    \\ \hline
  &              &    &     & 78.6  & 92.9          \\
\checkmark  &              &    &     & 82.9  & 94.3            \\
\checkmark  & \checkmark   &    &     & 83.9  & 94.6 \\
            &              & \checkmark&   & 82.5 & 93.4  \\
            &              & \checkmark & \checkmark   & 83.5   &  93.8  \\
\checkmark  &              & \checkmark &   & 84.0 &  94.6           \\
\checkmark  & \checkmark   & \checkmark & \checkmark & 84.6 & 95.0        \\ \hline
\end{tabular}}
\end{table}
\subsection{Comparison with State-of-the-art}

\textbf{Comparisons on MS-COCO.} As shown in \tabref{table: COCO benchmark}, we compare our approach on MS-COCO benchmark with 11 state-of-the-art methods.
The most commonly used resolution is 448$\times$448 during both training and testing stage. However, it is worthy to notice that some methods evaluate their performance on different resolutions during training and inference stages, \eg{}, ADD-GCN \cite{ye2020attention} and SSGRL \cite{chen2019learning}.
For fair comparisons, we follow their resolution settings \cite{chen2019multi,ye2020attention,chen2019learning} and report three results, which achieve a new performance leader-board with a clear margin.


\textbf{Comparisons on VOC 2007.} In \tabref{table: VOC2007 benchmark}, we compare our approach with 8 state-of-the-art methods.
For fair comparisons, we report mAP and AP of each class on commonly used 448$\times$448 resolution with only ImageNet pretrained. In terms of mAP, our approach achieves the best performance and outperforms state-of-the-art ML-GCN \cite{chen2019multi} by 1.0\%.

\subsection{Performance Analysis}

\begin{figure*}
	\begin{center}
		\includegraphics[width=1\textwidth]{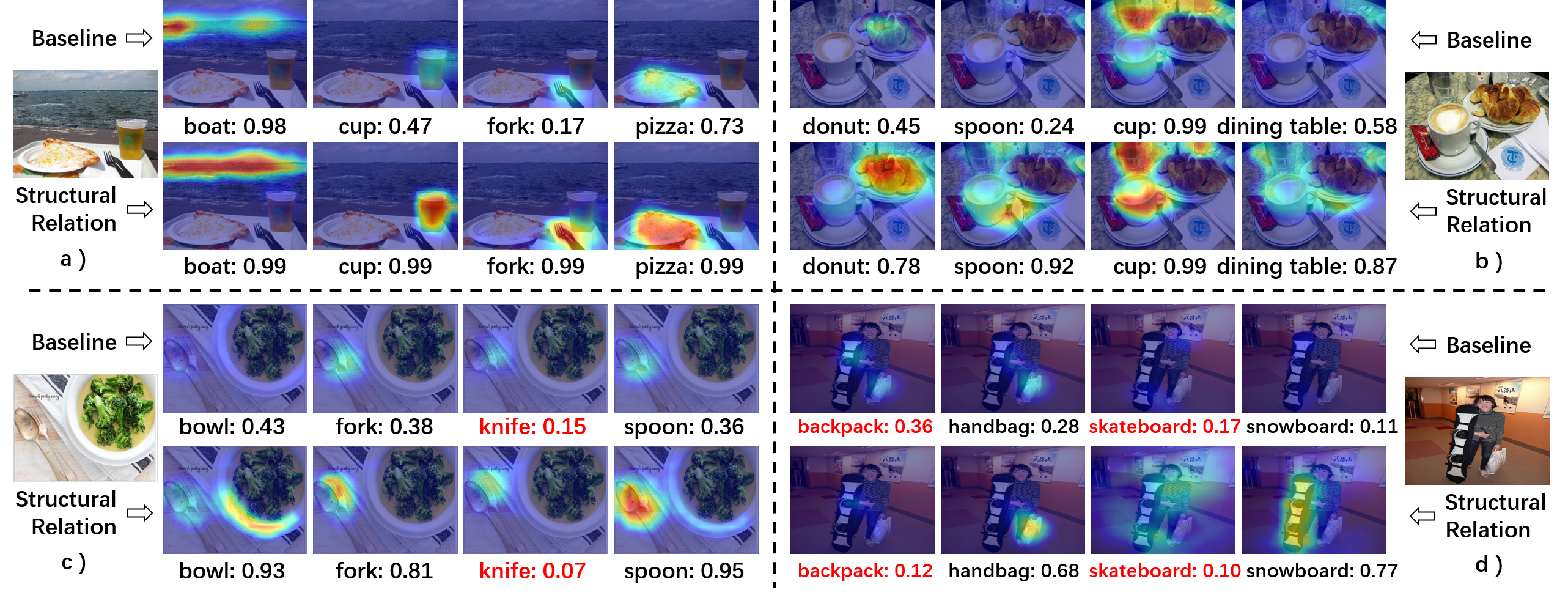}
		\caption{Visualization analyses of baseline and our proposed structural relation graph module. We present several labels for demonstration and the labels not presented in the image are highlighted in \textcolor{red}{red}. Compared to the baseline our structural relation has the ability to handle objects in small scales in a) and b) or with confused appearances in c) and d).
		}\label{fig:Qualitative results }
	\end{center}
 \vspace{-0.3cm}
\end{figure*}

\textbf{Ablation Studies.}
To evaluate the effectiveness of our proposed structural relation graph module and semantic relation graph module, we reconstruct our model with different ablation factors in \tabref{table:ablation}. We first employ ResNet-101 with the identical training protocol as our baseline model in the first row, which reaches a high baseline performance, \eg{}, 92.9\% on VOC 2007.
Note that this baseline model outperforms several state-of-the-art models and our proposed model can steadily improve the performance based on this high baseline.
As shown in \tabref{table:ablation}, our proposed modules make a steady improvement to the final performance, which demonstrates the necessity of the proposed modules to obtain the best classification results.


\begin{table}[]
\centering{
\caption{Ablation study for cross-scale attention module on MS-COCO. $\mc{S}_{\{\frac{1}{32},\frac{1}{64},\frac{1}{16}\}}$ denote features of different scales. $\mc{M}_{CA}$ denotes cross attention module. $\mc{M}_{Trans}$ denotes transformer units.}
\label{table:MCA ablation}
\begin{tabular}{ccccc|c}
\hline
\multirow{2}{*}{$\mc{S}_{\frac{1}{32}}$} & \multirow{2}{*}{$\mc{S}_{\frac{1}{64}}$} & \multirow{2}{*}{$\mc{S}_{\frac{1}{16}}$} & \multirow{2}{*}{$\mc{M}_{CA}$} & \multirow{2}{*}{$\mc{M}_{Trans}$} & \multicolumn{1}{c}{mAP} \\
  &   &   & &  & $\mc{R}_{Structural}$ \\ \hline
\checkmark &   &   &   & TR & 82.9                 \\
\checkmark & \checkmark &   &  & TR & 83.1  \textcolor{red}{($\uparrow$ 0.2)}       \\
\checkmark & \checkmark & \checkmark &  & TR & 83.3 \textcolor{red}{($\uparrow$ 0.4)} \\
\checkmark & \checkmark & \checkmark & SUM & TR & 83.2 \textcolor{red}{($\uparrow$ 0.3)}     \\
\checkmark & \checkmark & \checkmark & MUL & MLP & 83.3 \textcolor{red}{($\uparrow$ 0.4)}     \\
\checkmark & \checkmark & \checkmark & MUL & TR & 83.9 \textcolor{red}{($\uparrow$ 1.0)}     \\
\hline
\end{tabular}}
\end{table}

\textbf{Effects of Structural Relation.} It can be found in \tabref{table:ablation} that only adopting Transformer for structural relation can notably improve the performance, \eg{}, 4.3\% on MS-COCO, which demonstrates the effectiveness of long-term contextual information for multi-label recognition task.
Besides, the position information of small objects may vanish after down-sampling especially for challenging datasets, \eg{}, MS-COCO. Our proposed cross-scale attention module could enhance cross-scale features and suppress noises, which further boosts performance by 1.0\% on MS-COCO.

To verify the effectiveness of our proposed cross-scale attention module, we explore different scales on MS-COCO as shown in \tabref{table:MCA ablation}. The default scale is $\frac{1}{32}$ generated by the last stage of our baseline ResNet-101. With more different scales taken into consideration, the performance of structural relation graph module is slightly improved. With our proposed cross attention module with multiplication operation, the performance further boosts to a new level, the performance drops 0.7\% when replacing the multiplication with summation, which demonstrates our proposed cross attention module could further effectively strengthen the position information between different scales.

To further verify the effectiveness of transformer units on cross-scale information, we replace transformer units with a simple MLP layer in the 5th row in \tabref{table:MCA ablation}, the performance of $R_{structural}$ shows a clear drop (0.6\%) in mAP, which demonstrates the transformer units could effectively capture the long-term spatial context.


\begin{table}[]
\centering{
\caption{Ablation study for proposed joint relation graph on MS-COCO. $\mc{G}_{JCM}$ denotes the learnable joint correlation matrix. $\mc{G}_{SG}$ denotes structural guidance. $\mc{C}_{SAC}$ denotes the semantic-aware constraints. }
\label{table:GCN ablation}
\begin{tabular}{ccc|c}
\hline
\multirow{2}{*}{$\mc{G}_{JCM}$} & \multirow{2}{*}{$\mc{G}_{SG}$} & \multirow{2}{*}{$\mc{C}_{SAC}$} & \multicolumn{1}{c}{mAP} \\
 &  &  & $\mc{R}_{Semantic}$ \\ \hline
 Static &  &  & 81.9  \\
\checkmark &  &  & 82.5 \textcolor{red}{($\uparrow$ 0.6)}  \\
\checkmark & \checkmark &  & 83.0 \textcolor{red}{($\uparrow$ 1.1)}  \\
\checkmark & \checkmark & GMP  & 83.5 \textcolor{red}{($\uparrow$ 1.6)}   \\
\checkmark & \checkmark & GAP & 83.6 \textcolor{red}{($\uparrow$ 1.7)}  \\
\checkmark & \checkmark & KMP$_{5\%}$ & 83.7 \textcolor{red}{($\uparrow$ 1.8)}  \\
\hline
\end{tabular}}
\end{table}

\textbf{Effects of Semantic Relation.}
As shown in \tabref{table:ablation}, building semantic relationships with
GCN could notably improve the performance, \eg{} 3.9\% on MS-COCO. Moreover, the performance further boosts for 1\% with our proposed semantic-aware constraints, which demonstrates that GCN could achieve better modeling results with more representative semantic-specific vectors.
To evaluate the effectiveness of our proposed semantic relation graph module, we conduct detailed ablations on MS-COCO in \tabref{table:GCN ablation}. We adopt static adjacent matrix used in ML-GCN \cite{chen2019multi} to form a baseline in the first row. Applying our proposed learnable correlation matrix improves the performance by 0.6\%. Besides, joint semantic and structural information could effectively improve the performance by 0.5\% with structural guidance.
Another main exploration is to find global pooling operations for semantic-aware constraints, our final semantic relation achieves the best performance with top-k max-pooling operation with threshold 5\%.


\textbf{Interpretable Visualizations of Structural Relation.} We utilize Grad-CAM \cite{selvaraju2017grad} to exhibit the visualization results of baseline and the proposed structural relation in \figref{fig:Qualitative results }.
i) Benefiting from the cross-scale attention module, our approach could capture more accurate localization and effectively perceive \textbf{small objects}, \eg{} \textit{fork} in \figref{fig:Qualitative results } a) and \textit{spoon} in \figref{fig:Qualitative results } b).
ii) As baseline model could not distinguish objects with \textbf{similar appearances}, \eg{} the triplet labels \textit{\{fork, knife, spoon\}} in \figref{fig:Qualitative results } c) and the paired labels \textit{\{backpack, handbag\}, \{skateboard, snowboard\}} in \figref{fig:Qualitative results } d), these issues are well handled by our proposed structural relation module benefiting from the long-term contextual information captured by Transformer-based relation graph.

\section{Conclusion}
In this paper, we propose a novel Transformer-based Dual Relation Graph (TDRG) framework for multi-label recognition tasks.
We make the first attempt to introduce Transformer architecture into multi-label recognition tasks to incorporate long-term contextual information and build position-wise relationships across different scales.
Besides, we model dynamic co-occurrences with semantic-aware constraints. With these two complementary relations jointed,
our proposed approach achieves new state-of-the-art on two multi-label recognition benchmarks.

\section*{Acknowledgments}
This work was supported by grants from National Natural Science Foundation of China (No. 61922006).
\newpage
{\small
\bibliographystyle{ieee_fullname}
\bibliography{egbib}
}

\end{document}